\pdfoutput=1

\documentclass[11pt]{article}

\usepackage[]{acl}
\usepackage{multirow}

\usepackage{times}
\usepackage{latexsym}
\usepackage{makecell}

\usepackage[T1]{fontenc}

\usepackage[utf8]{inputenc}

\usepackage{microtype}
\usepackage{CJKutf8}        

\usepackage{inconsolata}
\usepackage{booktabs}
\usepackage{graphicx}

\usepackage{algorithm}
\usepackage{algpseudocode}
\usepackage{amsmath}

\usepackage{amsmath}
\usepackage{bm}
\usepackage{indentfirst}
\usepackage{subfigure}
\usepackage{epsfig}
\usepackage{epstopdf}
\usepackage{url}
\usepackage{xspace}
\usepackage{subeqnarray}
\usepackage{subcaption}
\usepackage{color}

\newcommand{\paratitle}[1]{\vspace{1.5ex}\noindent\textbf{#1}}
\newcommand{\ie}{\emph{i.e.,}\xspace}

\newcommand{\eg}{\emph{e.g.,}\xspace}

\newcommand{\ignore}[1]{}

\title{SimpleDeepSearcher: Deep Information Seeking via Web-Powered Reasoning Trajectory Synthesis}

\author{
    \textbf{Shuang Sun\textsuperscript{{1}}\thanks{\llap{}\:\:\:Equal contributions.},
            Huatong Song\textsuperscript{{1}}\footnotemark[1],
            Yuhao Wang\textsuperscript{{1}},
            Ruiyang Ren\textsuperscript{{1}},
            }\\
    \textbf{Jinhao Jiang\textsuperscript{{1}},
            Junjie Zhang\textsuperscript{{1}},
            Fei Bai\textsuperscript{{1}},
            Jia Deng\textsuperscript{{1}},
            }\\
    \textbf{Wayne Xin Zhao\textsuperscript{{1}}\thanks{\llap{}\:\:\:Corresponding authors.},
            Zheng Liu\textsuperscript{{2}},
            Lei Fang\textsuperscript{{3}}\footnotemark[2],
            Zhongyuan Wang\textsuperscript{{2}},
            Ji-Rong Wen\textsuperscript{{1}}}\\
	\textsuperscript{1}Gaoling School of Artificial Intelligence, Renmin University of China\\
	\textsuperscript{2}Beijing Academy of Artificial Intelligence\quad
        \textsuperscript{3}DataCanvas Alaya NeW\\
    \texttt{\{sunshuanguns, batmanfly\}@gmail.com}\\
}

\begin{document}
\maketitle
\begin{abstract}
Retrieval-augmented generation~(RAG) systems have advanced large language models~(LLMs) in complex deep search scenarios requiring multi-step reasoning and iterative information retrieval. 
However, existing approaches face critical limitations that lack high-quality training trajectories or suffer from the distributional mismatches in simulated environments and prohibitive computational costs for real-world deployment.
This paper introduces SimpleDeepSearcher, a lightweight yet effective framework that bridges this gap through strategic data engineering rather than complex training paradigms.
Our approach synthesizes high-quality training data by simulating realistic user interactions in live web search environments, coupled with a multi-criteria curation strategy that optimizes the diversity and quality of input and output side. Experiments on five benchmarks across diverse domains demonstrate that SFT on only 871 curated samples yields significant improvements over RL-based baselines.
Our work establishes SFT as a viable pathway by systematically addressing the data-scarce bottleneck, offering practical insights for efficient deep search systems.
Our code and data are available at \url{https://github.com/RUCAIBox/SimpleDeepSearcher}.
\end{abstract}

\section{Introduction}\label{sec:intro}

In recent years, retrieval-augmented generation~(RAG) methods have significantly enhanced LLMs by incorporating external knowledge retrieval mechanisms~\citep{lewis2020retrieval,rag_survey,gao2024retrievalaugmentedgenerationlargelanguage}. Recent advancements have extended these capabilities to complex \textit{deep search} scenarios that demand multi-step reasoning with iterative information retrieval and synthesis~\citep{alzubi2025open}. Traditional RAG systems typically treat retrieval as an external auxiliary module, following a fixed pipeline of ``retrieval–re-ranking–reading''~\citep{qi2020retrieve}. In contrast, deep search scenarios require the model to internalize the abilities of ``when to retrieve, how to retrieve, and how to reason based on retrieved content,'' in order to address more flexible and complex tasks.

To address the complex reasoning demands in deep search scenarios, early research explored prompt-based strategies that guide models to decompose questions, generate queries, and retrieve information iteratively~\citep{rag_star,teng2024atom,search_o1}.  Other studies have attempted to improve model performance through supervised fine-tuning (SFT)~\citep{cot_sft}; however, there is currently a lack of high-quality trajectory data of reasoning and search interactions for training~\citep{search-r1}.
To further enhance the model’s autonomous search capabilities, Reinforcement Learning (RL)~\citep{sutton1999reinforcement} is considered as a promising solution to train models through real-time interaction with the environment~\citep{nakano2021webgpt,r1-searcher,search-r1,zheng2025deepresearcher}.
However, most RL-based approaches operate within artificial environments using static document corpora, creating a distributional mismatch with real-world web dynamics. Moreover, the inherent computational intensity of RL training escalates exponentially when interfacing with live search APIs~\citep{sun2025zerosearch}. 

Given the overhead and complexity of RL-based training, we hypothesize that SFT remains a viable pathway for building efficient deep search systems. While SFT offers a streamlined training process, it faces the critical challenge of lacking high-quality training data in deep search scenarios. 
On the one hand, existing QA datasets often lack the diversity and complexity of questions and search-oriented purposes on the Web, which are essential for deep search training.
On the other hand, traditional answer annotations omit the critical reasoning traces~(search operations, evidence synthesis, and efficient decision paths) required for teaching search-integrated reasoning strategies.

In this paper, we propose SimpleDeepSearcher, an efficient search-with-think framework that utilizes strategic data engineering rather than complex training paradigms. Our core methodology centers on a three-fold process for constructing high-quality training data. First, we develop a data synthesis framework grounded in real web search environments, simulating realistic user search behaviors to generate multi-turn reasoning trajectories. Second, we propose a diversity-aware query sampling strategy to optimize domain coverage, semantic complexity, and knowledge unit density.
Moreover, we adopt a four-dimensional response curation that enforces format standardization, reasoning efficiency, question difficulty, and search effectiveness.
By systematically addressing both query and response-side quality through automated pipelines, SimpleDeepSearcher can obtain high-quality supervised signals based on real web search for complex reasoning to facilitate the SFT process.

Experimental results show that our SFT method significantly boosts model performance on five representative benchmarks with only 871 high-quality training samples. Compared to prompt-based methods, SimpleDeepSearcher achieves a 48.3\% improvement, and compared to RL-based RAG methods achieves a 24.9\% improvement. This demonstrates that our framework effectively balances performance and efficiency, providing a simple yet powerful approach to enhancing deep search capabilities. Furthermore, our framework is highly extensible that can be combined with other types of training data, the framework is also applicable to RL-based training.

Our main contributions are as follows:

$\bullet$ We propose a real web-based data synthesis framework that simulates realistic user search behaviors, generating multi-turn reasoning and search trajectories.

$\bullet$ We design a multi-criteria data curation strategy that jointly optimizes both input question selection and output response filtering through orthogonal filtering dimensions.

$\bullet$ Experimental results demonstrate that SFT on only 871 samples enables SimpleDeepSearcher to outperform strong baselines~(especially RL-based baselines) on both in-domain and out-of-domain benchmarks. 
\section{Method}\label{sec:method}








In this section, we propose SimpleDeepSearcher for complex deep search tasks by leveraging multi-stage data construction strategies.

\begin{figure*}[t]
    \centering
    \includegraphics[width=1\linewidth]{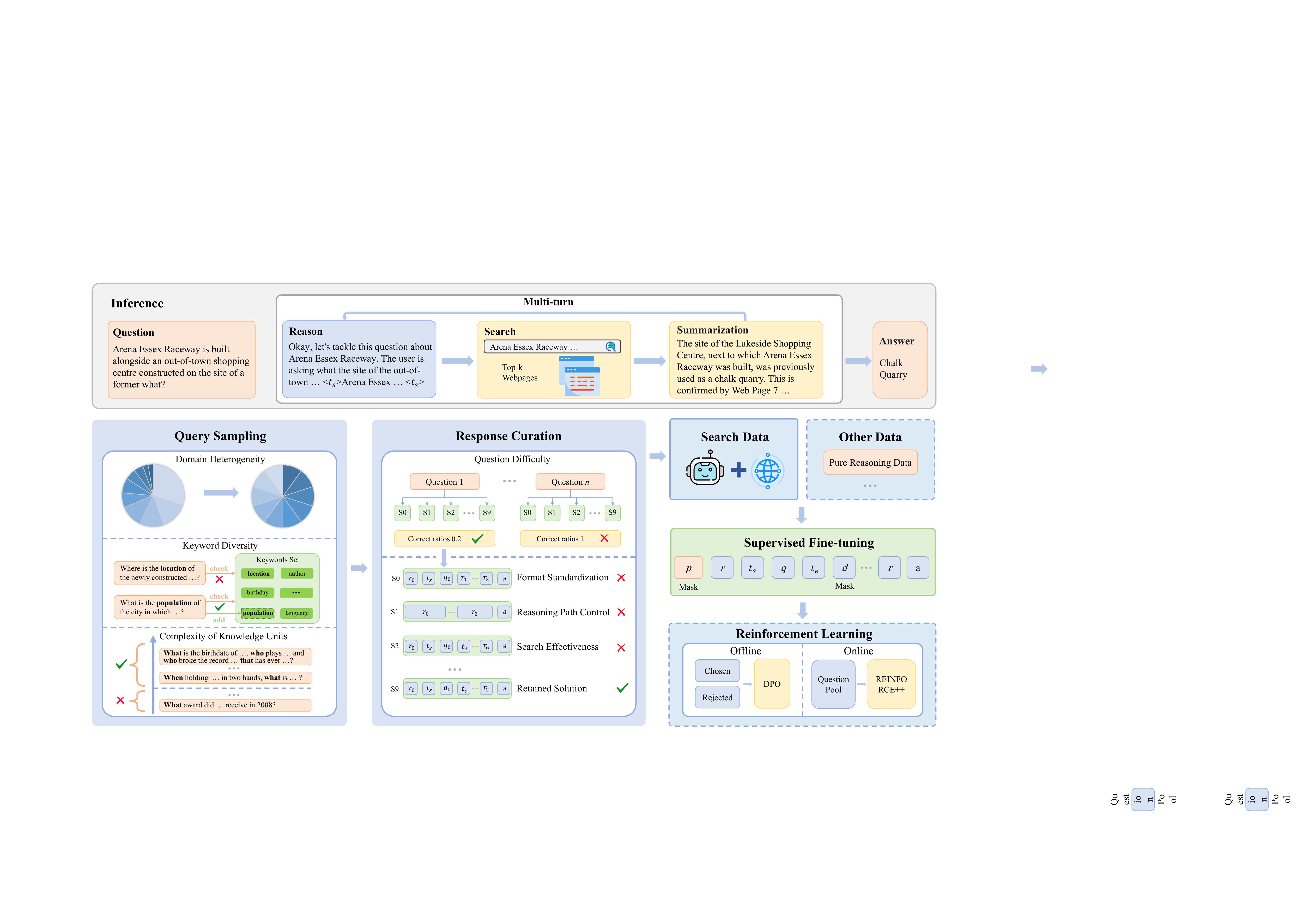}
    \caption{Overall framework of our proposed SimpleDeepSearcher approach. $r$ denotes the reasoning content, $q$ represents the search query, and $d$ refers to the retrieved document after summarization. $t_s$ and $t_e$ are special tokens indicating the beginning and end of the search query, and $a$ denotes the final answer.}
    \label{fig:overall_framework}
\end{figure*}

\subsection{Overview}

To address the resource-intensive limitations of deep search systems, we propose SimpleDeepSearcher, a framework that achieves intelligence search through efficient supervised fine-tuning (SFT) with minimal training data. For constructing high-quality SFT data, we establish a systematically designed data synthesis and curation pipeline, as illustrated in Figure~\ref{fig:overall_framework}.

First, we replace static document retrieval~\citep{karpukhin2020dense} with real-time network interactions, simulating human search behavior through an iterative cycle of "reasoning-searching-summarizing-generating." By directly processing raw HTML content via commercial search APIs, we capture diverse web information features—ranging from structured data snippets to unstructured narrative discourse. Based on this, we first filter input queries using domain heterogeneity, keyword diversity, and knowledge unit complexity to construct a maximally informative training foundation while ensuring selected queries align with real-world web search scenarios. Additionally, we apply a filtering mechanism to LLM-synthesized responses, implementing a four-dimensional quality filter that simultaneously optimizes format standardization, reasoning path control, question difficulty, and search effectiveness to guarantee response quality.

The framework's modular design offers three distinctive advantages: First, it exposes the model to authentic search artifacts and noise patterns through real web interactions. Second, our multidimensional filtering strategy enables state-of-the-art performance with remarkably small SFT datasets, eliminating dependency on resource-heavy RL training. Third, the decoupled architecture between data synthesis and model constraints provides exceptional flexibility that our curated datasets can enhance any LLMs while maintaining compatibility with emerging reasoning architectures and alternative training paradigms including RL.
Since the searched content is not generated by the LLM, we mask out these tokens during the SFT process.





Our methodology achieves unprecedented efficiency in search-oriented model training, reducing computational demands while maintaining competitive performance through strategic data quality optimization rather than brute-force data quantity.

\subsection{Data Synthesis in Real Web Environment}
\label{sec:data_synthesis}

Typically, traditional retrieval-augmented generation~(RAG) systems rely on closed and static knowledge corpora (\eg a Wikipedia snapshot). Such knowledge corpora exhibit two primary limitations: firstly, the content they contain often consists of refined and condensed segments~\citep{chen2024mindsearch}; secondly, the information within these knowledge corpora lacks timeliness. Consequently, RAG systems are limited in their ability to simulate authentic user search behaviors, as users typically search within open, dynamic, and complex web environments where the information is not only diverse in format and varied in quality but is also frequently accompanied by redundancy and noise. In light of this, our data synthesis approach does not rely on curated document collections but is instead grounded in the real, open web environment. This authentic web environment also places greater demands on the model's capabilities for information extraction, synthesis, and reasoning.

Building upon the widely adopted iterative deep search process~\citep{search_o1} of reason-search-summarize-generate, we develop an automated pipeline for large-scale training data synthesis. 
For each query, our framework at each iteration (1) initiates web searches via commercial APIs, (2) extracts and processes raw HTML content, (3) applies an LLM to reason over multi-source evidence, and (4) continues for the next iteration or stop iteration. By sampling multiple reasoning paths per query, we capture nuanced decision-making processes inherent to real-world information synthesis.

Our data synthesis strategy is firmly rooted in real web scenarios, which substantially enriches the diversity and authenticity of training samples. Building on this, this strategy enables scaling of high-quality trajectory data of reasoning and search interactions.

\subsection{Diversity-aware Query Sampling}
\label{sec:query_sampling}


To engineer a deep search architecture with advanced query comprehension and reasoning capabilities, we implement a strategic repurposing of open-domain question answering (QA) resources. These curated datasets offer natural language questions that inherently require multi-hop information retrieval operations, thereby exhibiting strong task alignment with the cognitive demands of deep search systems~\citep{zheng2025deepresearcher}. Our selection protocol combines single-hop and multi-hop QA benchmarks through principled composition, ensuring coverage of both atomic and composite reasoning paradigms.



However, empirical evidence suggests that naive dataset scaling yields diminishing returns in SFT~\citep{zhou2023lima}. The efficacy of such approaches fundamentally depends on the intrinsic diversity and informational entropy of training instances. While existing open-domain QA corpora provide substantial volume, systematic analysis reveals three critical limitations: (1) domain-specific overrepresentation creating skewed knowledge distributions, (2) repetitive syntactic patterns reducing linguistic variability~\citep{parmar2022don}, and (3) semantic simplicity thresholds below real-world query complexity. These factors collectively induce model brittleness and constrain cross-domain generalization potential (see Appendix~\ref{sec:daqs} for details).
To address these critical limitations, we introduce a diversity-aware query sampling strategy to implement systematic data filtering through tripartite orthogonal criteria:

\paratitle{Domain Heterogeneity} encompasses the systematic classification of query semantics across distinct knowledge domains (\eg history, science, politics). This dimension ensures a balanced distribution of questions across different domains, thereby reducing domain-specific biases and enhancing generalization capabilities.

\paratitle{Keyword Diversity} 
focuses on the distributional diversity of core semantic constituents (definition provided in Appendix~\ref{sec:daqs}).
we ensure non-redundant exposure to low-frequency conceptual entities, multi-order relational dependencies, and contextually ambiguous referential expressions. Such systematic variation compels the model to transcend superficial lexical pattern matching, instead developing reasoning architectures essential for interpreting complex entity interactions~\citep{linzen2020can}.

\paratitle{Complexity of knowledge units} captures the frequency of interrogative terms used in questions (\eg what, when), which serve as indicators of syntactic and semantic complexity. 
Questions with greater inquiry potential are given priority, ensuring comprehensive modeling of implicit reasoning chains triggered by diverse question formulations. 


We developed a systematic query selection framework incorporating three complementary dimensions: domain heterogeneity, keyword diversity, and complexity of knowledge units. 
First, we partition the dataset into domain-specific clusters using the LLM-generated semantic classifications. Within each domain cluster, queries are ranked by knowledge unit complexity scores derived from conceptual density analysis. Subsequently, we perform iterative selection using a greedy algorithm that maximizes keyword diversity while maintaining inter-domain balance. 
The detailed procedure for query sampling is presented in Algorithm~\ref{alg:data_selection}. 

\begin{algorithm}[t]
\small
\caption{Diversity-aware Query Sampling}
\label{alg:data_selection}
\begin{algorithmic}[1]
\Require \raggedright Annotated dataset $D$ with domains $d_1, d_2, \dots, d_m$, target number of queries $N$
\State $N_d \gets N / m$ 
\State $S \gets \emptyset$ \Comment{Initialize the target set}
\For{$i$ = 1 to $n$}
    \State $D_{d_i} \gets \{x \in D \mid \text{domain}(x) = d_i\}$ 
    \State Sort $D_{d_i}$ by descending interrogative 
    \State words
    \While{$|S_{d_i}| < N_d$ and $D_{d_i} \neq \emptyset$}
        \State $K \gets \emptyset$ \Comment{Initialize the keyword set}
        \For{each sample $x$ in $D_{d_i}$}
            \If{$|S_{d_i}| \geq N_d$}
                \State \textbf{break}
            \EndIf
            \State $kw \gets \text{keywords}(x)$
            \If{$x \notin S$ and $kw \cap K = \emptyset$}
                \State $S \gets S \cup \{x\}$
                \State $K \gets K \cup \text{keywords}(x)$
                \State $D_{d_i} \gets D_{d_i} \setminus \{x\}$
            \EndIf
        \EndFor
    \EndWhile
\EndFor
\State \Return $S$
\end{algorithmic}
\end{algorithm}

\subsection{Multi-Dimention Response Curation}
\label{sec:response_curation}






Building upon the aforementioned data synthesis and query sampling strategies, we have successfully generated high-quality training data derived from real-world web environments. However, due to the inherent unpredictability of LLM reasoning, the quality of synthesized data exhibits considerable variability despite meticulous control over input and generation processes. 
Three primary issues are observed: (i) Formatting irregularities, such as inconsistent reasoning languages, non-standard formats for search and reasoning steps, and heterogeneous answer formats; (ii) reasoning redundancy, including hypothesis overgeneration, fabricated retrieval content, and excessive validation loops; (iii) inefficient search strategies, including redundant search exploration, contextual myopia and failure to retrieve relevant information.

The presence of low-quality reasoning outputs in language models not only compromises performance and transparency but also introduces noise into training signals, leading to inefficient computational resource utilization. To address these challenges, we developed a systematic filtering protocol that selects optimal solutions through rigorous evaluation of multiple responses per query. 


To mitigate these issues, we impose strict constraints on both the format and content of sampled responses, retaining only those that satisfy all predefined criteria. Our filtering strategy, structured around four pillars, ensures retention of high-quality reasoning data while promoting efficient search integration.

\paratitle{Format Standardization}. Filter out responses with mixed reasoning languages or incorrect reasoning and search formats, and correct answers with formatting errors to ensure consistency and standardization across all responses.
Responses exhibiting mixed languages, irregular reasoning structures, or formatting inconsistencies were excluded. Automated correction aligned remaining answers with standardized templates.

\paratitle{Reasoning Path Control}. Strictly limit the use of reflection expressions (\eg alternatively, wait, etc.) and control the length of reasoning to avoid unnecessary and redundant reasoning steps. Reasoning models tend to hypothesize, infer, and reflect based on internal knowledge, often resulting in delayed use of search tools and inefficient reasoning. By regulating the reasoning path, the model can learn to seamlessly integrate search into its inference process and adopt more efficient reasoning strategies.

\paratitle{Question Difficulty}. Filter out questions with consistently high accuracy across multiple reasoning attempts and prioritize those with lower accuracy. Accuracy obtained from multiple samples can serve as a proxy for question difficulty. Selecting more challenging questions helps enhance the model's ability to handle complex queries.

\paratitle{Search Effectiveness}. Among multiple candidate responses, prioritize those with fewer search steps and more diverse search content. This encourages the model to not only invoke search capabilities but also to learn how to formulate effective sub-queries based on the original question for efficient information retrieval.

Based on the above dimensions, we first collect metadata for each response, such as the number of search steps, reasoning length, and accuracy. Subsequently, responses are filtered sequentially based on \textit{format standardization} and \textit{reasoning path control}. Then, based on \textit{question difficulty}, questions with high accuracy are removed. For each remaining question, we retain multiple high-quality responses that meet all constraints and sort them by search steps. According to \textit{search effectiveness}, the response with the fewest search steps is selected as the final answer. Through this process, we ultimately obtained 871 high-quality question-answer pairs.
This multi-criteria approach not only enhances model training efficiency but also provides insights into optimal human-AI reasoning patterns.

\begin{table*}[t]
\centering
\small
\scalebox{0.93}{
\begin{tabular}{llcccccccccc}
\toprule 
\multirow{2}{*}{\textbf{Models}} & \multirow{2}{*}{\textbf{Methods}} & \multicolumn{2}{c}{\textbf{2Wiki$^\dagger$}} & \multicolumn{2}{c}{\textbf{MuSiQue$^\dagger$}} & \multicolumn{2}{c}{\textbf{Bamboogle$^\ddagger$}} & \multicolumn{2}{c}{\textbf{Frames$^\ddagger$}} & \multicolumn{2}{c}{\textbf{GAIA$^\ddagger$}} \\
\cmidrule(lr){3-4} \cmidrule(lr){5-6} \cmidrule(lr){7-8} \cmidrule(lr){9-10} \cmidrule(lr){11-12} 
 & & F1 & LasJ & F1 & LasJ & F1 & LasJ & F1 & LasJ   & F1 & LasJ  \\
\midrule

 \multirow{6}{*}{\textbf{Qwen-7B}} 
 & Directly Gen  & 27.7\hspace*{1.5mm} & 26.8\hspace*{1.5mm} & 9.6\hspace*{1.5mm} & 6.2\hspace*{1.5mm} & 18.2\hspace*{1.5mm} & 17.6\hspace*{1.5mm} & 12.6\hspace*{1.5mm} & 10.1\hspace*{1.5mm} & 13.6\hspace*{1.5mm} & 6.8\hspace*{1.5mm} \\
& Standard RAG  & 34.8\hspace*{1.5mm} & 34.8\hspace*{1.5mm} & 17.2\hspace*{1.5mm} & 14.6\hspace*{1.5mm} & 31.5\hspace*{1.5mm} & 31.2\hspace*{1.5mm} & 13.9\hspace*{1.5mm} & 13.5\hspace*{1.5mm} & - & - \\
 & Search-o1  & 48.0\hspace*{1.5mm} & 51.2\hspace*{1.5mm} & 21.5\hspace*{1.5mm} & 20.6\hspace*{1.5mm} & 57.9\hspace*{1.5mm} & 59.2\hspace*{1.5mm} & 30.9\hspace*{1.5mm} & 35.0\hspace*{1.5mm} & \underline{24.3}\hspace*{1.5mm} & \underline{21.4}\hspace*{1.5mm} \\
 & R1-Searcher & \underline{63.4}\hspace*{1.5mm} & 66.4\hspace*{1.5mm} & \textbf{29.0}\hspace*{1.5mm} & 26.8\hspace*{1.5mm} & 68.2\hspace*{1.5mm} & 68.8\hspace*{1.5mm} & \underline{34.4}\hspace*{1.5mm} & \underline{40.3}\hspace*{1.5mm} & 24.1\hspace*{1.5mm} & 20.4\hspace*{1.5mm} \\
 & DeepResearcher & 59.7$^*$ & \underline{66.6}$^*$ & 27.1$^*$ & \underline{29.3}$^*$ & \underline{71.0}$^*$ & \underline{72.8}$^*$ & - & - & - & - \\
 & SimpleDeepSearcher  & \textbf{70.6}\hspace*{1.5mm} & \textbf{79.8}\hspace*{1.5mm} & \underline{28.2}\hspace*{1.5mm} & \textbf{29.4}\hspace*{1.5mm} & \textbf{74.5}\hspace*{1.5mm} & \textbf{76.8}\hspace*{1.5mm} & \textbf{44.9}\hspace*{1.5mm} & \textbf{55.3}\hspace*{1.5mm} & \textbf{39.3}\hspace*{1.5mm} & \textbf{36.9}\hspace*{1.5mm} \\
 
\cmidrule{2-2}

\multirow{4}{*}{\textbf{Qwen-32B}} 
 & Directly Gen & 31.7\hspace*{1.5mm} & 31.2\hspace*{1.5mm} & 13.3\hspace*{1.5mm} & 12.4\hspace*{1.5mm} & 25.7\hspace*{1.5mm} & 25.6\hspace*{1.5mm} & 15.6\hspace*{1.5mm} & 14.2\hspace*{1.5mm} & 18.6\hspace*{1.5mm} & 13.9\hspace*{1.5mm} \\
& Standard RAG & 43.7\hspace*{1.5mm} & 45.0\hspace*{1.5mm} & 19.5\hspace*{1.5mm} & 16.8\hspace*{1.5mm} & 40.8\hspace*{1.5mm} & 40.8\hspace*{1.5mm} & 19.4\hspace*{1.5mm} & 19.4\hspace*{1.5mm} & - & - \\
 & Search-o1 & 64.9\hspace*{1.5mm} & \underline{74.8}\hspace*{1.5mm} & \underline{29.1}\hspace*{1.5mm} & \underline{30.6}\hspace*{1.5mm} & \underline{74.4}\hspace*{1.5mm} & \underline{78.4}\hspace*{1.5mm} & \underline{47.2}\hspace*{1.5mm} & \underline{56.8}\hspace*{1.5mm} & \underline{36.5}\hspace*{1.5mm} & \underline{34.0}\hspace*{1.5mm} \\
 & SimpleDeepSearcher & \textbf{71.9}\hspace*{1.5mm} & \textbf{81.2}\hspace*{1.5mm} & \textbf{30.6}\hspace*{1.5mm} & \textbf{33.0}\hspace*{1.5mm} & \textbf{78.1}\hspace*{1.5mm} & \textbf{80.0}\hspace*{1.5mm} & \textbf{50.1}\hspace*{1.5mm} & \textbf{60.8}\hspace*{1.5mm} & \textbf{42.1}\hspace*{1.5mm} & \textbf{40.8}\hspace*{1.5mm} \\
 
\cmidrule{2-2}
\multirow{4}{*}{\textbf{DDQ-32B }} 
& Directly Gen & 36.9\hspace*{1.5mm} & 36.2\hspace*{1.5mm} & 19.6\hspace*{1.5mm} & 16.0\hspace*{1.5mm} & 32.6\hspace*{1.5mm} & 32.8\hspace*{1.5mm} & 27.8\hspace*{1.5mm} & 29.2\hspace*{1.5mm} & 14.8\hspace*{1.5mm} & 9.7\hspace*{1.5mm} \\
& Standard RAG  & 48.1\hspace*{1.5mm} & 50.0\hspace*{1.5mm} & 24.0\hspace*{1.5mm} & 21.6\hspace*{1.5mm} & 42.6\hspace*{1.5mm} & 46.4\hspace*{1.5mm} & 26.5\hspace*{1.5mm} & 28.9\hspace*{1.5mm} & - & - \\
& Search-o1  & \underline{49.6}\hspace*{1.5mm} & \underline{55.2}\hspace*{1.5mm} & \underline{25.4}\hspace*{1.5mm} & \underline{23.8}\hspace*{1.5mm} & \underline{65.7}\hspace*{1.5mm} & \underline{68.0}\hspace*{1.5mm} & \underline{32.2}\hspace*{1.5mm} & \underline{38.7}\hspace*{1.5mm} & \underline{23.2}\hspace*{1.5mm} & \underline{24.3}\hspace*{1.5mm} \\
& SimpleDeepSearcher  & \textbf{69.0}\hspace*{1.5mm} & \textbf{77.4}\hspace*{1.5mm} & \textbf{32.9}\hspace*{1.5mm} & \textbf{33.6}\hspace*{1.5mm} & \textbf{80.5}\hspace*{1.5mm} & \textbf{83.2}\hspace*{1.5mm} & \textbf{52.2}\hspace*{1.5mm} & \textbf{63.8}\hspace*{1.5mm} & \textbf{42.0}\hspace*{1.5mm} & \textbf{41.7}\hspace*{1.5mm} \\

\cmidrule{2-2}
\multirow{4}{*}{\textbf{QwQ-32B}} 
 
& Directly Gen & 39.6\hspace*{1.5mm} & 39.8\hspace*{1.5mm} & 18.9\hspace*{1.5mm} & 17.4\hspace*{1.5mm} & 29.6\hspace*{1.5mm} & 29.6\hspace*{1.5mm} & 28.1\hspace*{1.5mm} & 31.3\hspace*{1.5mm} & 16.8\hspace*{1.5mm} & 11.7\hspace*{1.5mm} \\ 
& Standard RAG  & 48.4\hspace*{1.5mm} & 50.6\hspace*{1.5mm} & 21.8\hspace*{1.5mm} & 19.4\hspace*{1.5mm} & 42.5\hspace*{1.5mm} & 46.4\hspace*{1.5mm} & 27.4\hspace*{1.5mm} & 31.6\hspace*{1.5mm} & - & - \\
& Search-o1  & \underline{69.4}\hspace*{1.5mm} & \underline{78.0}\hspace*{1.5mm} & \underline{34.3}\hspace*{1.5mm} & \underline{36.4}\hspace*{1.5mm} & \underline{78.7}\hspace*{1.5mm} & \underline{78.4}\hspace*{1.5mm} & \underline{51.6}\hspace*{1.5mm} & \underline{64.4}\hspace*{1.5mm} & \underline{38.3}\hspace*{1.5mm} & \underline{37.9}\hspace*{1.5mm} \\
& SimpleDeepSearcher  & \textbf{75.6}\hspace*{1.5mm} & \textbf{84.4}\hspace*{1.5mm} & \textbf{34.8}\hspace*{1.5mm} & \textbf{37.4}\hspace*{1.5mm} & \textbf{83.4}\hspace*{1.5mm} & \textbf{88.0}\hspace*{1.5mm} & \textbf{56.8}\hspace*{1.5mm} & \textbf{68.8}\hspace*{1.5mm} & \textbf{48.9}\hspace*{1.5mm} & \textbf{50.5}\hspace*{1.5mm} \\

\bottomrule
\end{tabular}
}
\caption{Performance comparisons between SimpleDeepSearcher and the baselines on QA benchmarks. The best results are in \textbf{bold} and the second-best are \underline{underlined}. $^\dagger/\ddagger$ represents in-domain/out-domain datasets. Results marked with * are cited from their official paper or report. \textit{Qwen-7B}, \textit{Qwen-32B}, \textit{DDQ-32B} are the abbreviations of Qwen-2.5-7B-Instruct, Qwen-2.5-32B-Instruct, and Deepseek-Distilled-Qwen-2.5-32B, respectively.}
\label{tab:main_results} 
\end{table*}

\begin{table}[t]
    \centering
    \small
    \resizebox{1\linewidth}{!}{
        \begin{tabular}{l c c c}
            \toprule
            \textbf{Model} 
            & \textbf{Xbench-DeepSearch}  
            & \textbf{BrowseComp-ZH}
            & \textbf{BrowseComp-EN} \\
            \midrule
            Webthink-RL     & 24.0$^*$ & 7.3$^*$ & 2.8$^*$ \\
            WebDancer-32B  & \textbf{38.7$^*$} & 14.1$^*$ & 2.5$^*$ \\
            SimpleDeepSearcher & 30.0\hspace*{1.5mm} & \textbf{14.5}\hspace*{1.5mm} & \textbf{4.3}\hspace*{1.5mm} \\
            \bottomrule
        \end{tabular}
    }
    \caption{Results on the more challenging Xbench-DeepSearch, BrowserComp-ZH, and BrowseComp-EN benchmarks. The results are evaluated with LLM-as-Judge. Results marked with * are cited from other papers or reports. The best results are in \textbf{bold}.}
    \label{tab:extra_benchmarks}
\end{table}

\section{Experiments}\label{sec:exp}
\subsection{Experimental Setup}



\paragraph{Datasets.} 
We sample training data from single-hop and multi-hop knowledge-intensive QA datasets to cover a wide range of domains and question difficulty. For single-hop questions, we use Natural Questions~\citep{nq} and SimpleQA~\citep{simpleqa}. For multi-hop questions, we use HotpotQA~\citep{yang2018hotpotqa}, 2WikiMultiHopQA~\citep{2wiki}, MuSiQue~\citep{tang2024multihop}, and MultiHopRAG~\citep{tang2024multihop}. To test the model’s performance on out-of-domain data, we select Bamboogle~\citep{bamboogle}, FRAMES~\citep{frames}, and GAIA~\citep{mialon2023gaia} benchmarks. In addition, we further conduct evaluations on the more challenging benchmarks including xbench-DeepSearch, BrowseComp-ZH~\citep{zhou2025browsecomp}, and BrowseComp-EN~\citep{wei2025browsecomp}. These datasets are not used during training and help evaluate how well the model works on new domains.
We evaluate on 500 randomly sampled instances from the validation sets of HotpotQA, 2WikiMultiHopQA, and MuSiQue. For GAIA, we use 103 examples from the text-only validation subset~\citep{li2025webthinker}, while for BrowseComp-EN we randomly sample 300 instances. For the remaining benchmarks, we use their full test sets.

\paragraph{Metrics.} 
We report results using two metrics: F1 score and LLM-as-Judge (LasJ). The F1 score captures the word-level similarity between the predicted and golden answers, while LasJ leverages GPT-4o-mini to evaluate the correctness of the predicted response.



\paragraph{Baselines.}


We consider following type of baselines:
\textit{Naive Generation}: Direct generation of answers without retrieval.
\textit{Standard RAG}\citep{rag_survey}: Directly retrieve relevant documents by querying the original question.
\textit{Search-o1}~\citep{search_o1}: Encourages the model to perform self-initiated retrieval using prompts.
\textit{RAG-RL}: R1-Searcher~\citep{r1-searcher}, DeepResearcher~\citep{zheng2025deepresearcher}, WebThinker-RL~\citep{li2025webthinker}, and WebDancer~\citep{wu2025webdancer}, the open-source models trained with reinforcement learning to enable self-initiated retrieval.
We conduct experiments using the following model backbones with an online search engine, including Qwen-2.5-7B-Instruct, Qwen-2.5-32B-Instruct, Deepseek-Distilled-Qwen-2.5-32B, and QwQ-32B.

\paragraph{Implementation Details.} 

Our experimental setup consists of three main components: SFT, generation, and query sampling. In the SFT phase, we use a total batch size of 64 and train for 6 epochs with a learning rate of 1e‑5, warmup ratio of 0.03, and a sequence length of 30,000 tokens. During fine-tuning, external retrieval documents are masked to avoid learning from noisy or spurious information.
For generation, all models are configured with a maximum sequence length of 20,480 tokens, temperature of 0.6, top‑p of 0.95, and top‑k of 40. 
During query sampling, we used QwQ-32B to annotate each query with its corresponding domain and keywords. For data synthesis, we employed QwQ-32B as the reasoning model and Google Search API as the search engine, with a maximum of 10 search calls and 15 reasoning turns per query. For each query, we sampled 10 candidate responses. All prompts used in the experiments are provided in Appendix~\ref{sec:instrcution_templates}.




\begin{table}[t]
\centering
\small
\resizebox{1\linewidth}{!}{
\begin{tabular}{l c c c  c c }
    \toprule
    \multirow{2.5}{*}{\textbf{Category}} & \multirow{2.5}{*}{\textbf{Method}} & \multicolumn{2}{c}{\textbf{Bamboogle}} & \multicolumn{2}{c}{\textbf{GAIA}}  \\
    \cmidrule(r){3-4}\cmidrule(r){5-6}
   &  & \textbf{F1} & \textbf{LasJ} & \textbf{F1} & \textbf{LasJ} \\
    \midrule
    & Ours & \textbf{74.5} & \textbf{76.8} & \textbf{39.3} & \textbf{36.9}  \\
    \midrule
    \multirow{3}{*}{{Query Sampling}}& w/o DD  & 69.7 & 70.4 & 35.6 & 35.8  \\
    & w/o KD  & 73.2 & 76.0 & 32.9 & 31.1  \\
    & w/o CIW  & 71.7 & 74.4 & 32.1 & 29.1  \\
    \midrule
    \multirow{1}{*}{{Environment}}& w/o Online &74.0&  74.4&  30.4 &28.2 \\
    \midrule
     \multirow{4}{*}{{Response Curation}}& w/o FS    & 72.8 & 75.2 & 38.0 & \textbf{36.9} \\
    & w/o RPC   & 71.7 & 74.4 & 31.6 & 30.1  \\
    & w/o QD    & 67.1 & 70.4 & 32.9 & 32.0  \\
    & w/o SE    & 72.6 & 73.6 & 37.7 & 35.0  \\
    
    \bottomrule
\end{tabular}
}
\label{tab:ablation_query}
\caption{Results of variants of SimpleDeepSearcher on Bamboogle and GAIA.}
\end{table}

\subsection{Main Results}

Table~\ref{tab:main_results} presents the main results of the proposed SimpleDeepSearcher and baselines across five representative datasets.


Firstly, SimpleDeepSearcher consistently outperforms all existing baseline methods across five benchmark datasets. Specifically, it achieves the best performance not only on in-domain datasets (\ie 2Wiki, MuSiQue) but also shows substantial improvements on out-of-domain datasets (\ie Bamboogle, FRAMES, GAIA), demonstrating its strong generalization ability.

Besides, SimpleDeepSearcher consistently outperforms reinforcement learning-based methods such as R1-Searcher and DeepResearcher across most evaluation metrics. These approaches are trained on large-scale datasets using complex reinforcement learning algorithms. In contrast, our method relies on supervised fine-tuning with only 871 training examples. This demonstrates that our framework achieves strong performance while maintaining high data efficiency, offering a simple yet effective alternative for improving deep search capabilities.


Thirdly, SimpleDeepSearcher achieves stable and substantial performance improvements across models with diverse backbones and parameter scales, ranging from 7B to 32B. For instance, compared to Search-o1, it achieves relative improvements of 48.3\%, 42.6\%, and 11.5\% on Qwen2.5-7B-Instruct, DeepSeek-R1-Distill-Qwen-2.5-32B, and QwQ-32B, respectively. This demonstrates the strong generalization ability of our distillation and self-distillation strategies, with the selected data consistently leading to performance gains across heterogeneous model architectures.


In addition, table~\ref{tab:extra_benchmarks} presents the experimental results on more complex QA datasets. These datasets are specifically designed for AI agents, requiring models to possess end-to-end planning, search, reasoning, and summarization capabilities. Notably, our model still demonstrates strong performance compared to models trained with reinforcement learning. This result further underscores the robust generalization ability of our model.
\section{Further Analysis}

\subsection{Ablation Study}

To validate the effectiveness of the proposed SimpleDeepSearcher, we conduct a comprehensive ablation analysis using Qwen2.5-7B-Instruct on the Bamboogle and GAIA datasets.
We conduct detailed ablation studies on three main aspects: (1) Query Sampling: \textit{w/o DD} removes domain diversity filter, \textit{w/o KD} removes keyword diversity filter, \textit{w/o CIW} removes coverage of 
interrogative words filter; (2) Environment: \textit{w/o Online} uses local dense dense retrieval to synthesize training data; (3) Response Curation: \textit{w/o FR} removes 
format regularization filter, \textit{w/o RPC} removes reasoning path control filter, \textit{w/o QD} removes question difficulty filter, \textit{w/o SC}  search count filter. As observed, all ablated variants exhibit a decline in performance compared to our full method, underscoring the integral contribution of each component. Among them, \textit{w/o QD} leads to the most significant performance drop, suggesting that question difficulty plays a crucial role in training. More challenging questions are more likely to stimulate the model’s autonomous retrieval capabilities during reasoning.

\begin{table}[t]
\centering
\small
\begin{tabular}{l  c c  c c   }
    \toprule
    \multirow{2.5}{*}{\textbf{Method}} & \multicolumn{2}{c}{\textbf{Bamboogle}} & \multicolumn{2}{c}{\textbf{GAIA}}   \\
    \cmidrule(r){2-3}\cmidrule(r){4-5}
    & \textbf{F1} & \textbf{LasJ} & \textbf{F1} & \textbf{LasJ}  \\
    \midrule
    Distilled (Ours)& 74.5 & 76.8 & \textbf{39.3} & 36.9  \\
    \midrule
    w. DPO    & \textbf{75.0} & \textbf{79.2} & 39.0 & \textbf{37.9}   \\
    w. Reinforce++   & 73.8& 75.8 & 29.4  &	24.3   \\

    \bottomrule
\end{tabular}

\caption{Evaluation Results of RL-based Methods.}
\label{tab:ablation_rl}
\end{table}

\begin{table}[t]
    \centering
    \small
    \resizebox{1\linewidth}{!}{
        \begin{tabular}{l c c c}
            \toprule
            \textbf{Model} 
            & \textbf{\#Alternatively}  
            & \textbf{\#Search}
            & \textbf{Output Length} \\
            \midrule
            QwQ-32B     & 7.933 & 2.390 & 867.148 \\
            QwQ-32B-SFT & 4.051 & 2.329 & 581.731 \\
            \bottomrule
        \end{tabular}
    }
    \caption{Statistical analysis of model outputs.}
    \label{tab:qwq32b_sft_output_stats}
\end{table}

\subsection{Effect of Post-SFT RL}

Recent studies have investigated the integration of RL and RAG~\citep{r1-searcher, search-r1, zheng2025deepresearcher}. We further examine the advantages and limitations of applying RL after SFT.

We apply DPO and REINFORCE++ to conduct offline and online reinforcement learning, respectively. As shown in Table~\ref{tab:ablation_rl}, the model trained with DPO achieves further improvements over the SFT baseline, demonstrating the effectiveness of offline preference optimization (see Appendix~\ref{sec:dpo_setting} for details).
In contrast, the model trained with REINFORCE++ produces significantly shorter responses~(see Appendix~\ref{sec:appendix-on-policy} for details) and shows notable performance degradation on both the Bamboogle and GAIA benchmarks. This suggests that online RL mainly triggers retrieval behavior, but brings little benefit to models that are already good at retrieval.
We hypothesize that the success of offline DPO stems from its ability to leverage high-quality trajectories generated by a strong LLM. These trajectories provide informative preference signals and stable supervision, allowing the model to refine its reasoning and search strategies.

\begin{table}[t]
\small
\centering
{
\begin{tabular}{llccc}
\toprule
\textbf{Model} & \textbf{Plan.} & \textbf{Search} & \textbf{Summ.} \\
\midrule
Qwen-7B      & 0.416 & 0.455 & 0.363 \\
Qwen-7B-SFT  & \textbf{0.590} & \textbf{0.677} & \textbf{0.584} \\
\midrule
QwQ-32B                 & 0.623 & 0.680 & 0.594 \\
QwQ-32B-SFT             & \textbf{0.629} & \textbf{0.713} & \textbf{0.624} \\
\bottomrule
\end{tabular}
}
\caption{Proportion of instances containing the correct answer at each stage of the inference process (Planning, Search, and Summarization), before and after SFT.}
\label{tab:search_capability}
\end{table}

\subsection{Effect of SFT on Redundancy}


In this part, we analyze how SFT impacts redundant reasoning and search behavior. Specifically, we focus on three indicators: (1) the frequency of the reflective word ``alternatively'', which signals hesitation or divergent reasoning; (2) the average length of reasoning chains, measured by output length; and (3) the number of search calls made during inference.
Our analysis is based on the QwQ-32B model, evaluated on the 2Wiki, MuSiQue, and Bamboogle datasets.
As shown in Table~\ref{tab:qwq32b_sft_output_stats}, the average use of ``alternatively'' and the overall output length are both significantly reduced after SFT. Moreover, the model issues fewer search queries. These results indicate that our self-distillation approach improves both the reasoning clarity and search efficiency of the model. This improvement can be attributed to the high-quality training data selected through our proposed method.


\subsection{Effect of SFT on Stage-wise Performance}



In this part, we analyze how training improves the performance of each sub-task in our approach, including iterative search, planning, and summarization.
We evaluate the proportion of cases in which the final answer appears during each sub-process to quantify the efficiency of that stage.
To eliminate interference from the summarization stage, all summarization models are kept identical during inference, with detailed settings provided in Appendix \ref{sec:model_enhancement}. The results are shown in Table ~\ref{tab:search_capability}. We can observe substantial improvements across all components, with the search component showing the most significant gain. This suggests that training effectively enhances the model's ability to generate more coherent reasoning and search trajectories, leading to more accurate information retrieval and improved overall model performance.

\begin{table}[t]
\centering
\resizebox{1\linewidth}{!}{
\begin{tabular}{lccccc}
\toprule
\multirow{2}{*}{\textbf{Models}}
& \multirow{2}{*}{\textbf{\makecell{Summarization Model}}}
& \multicolumn{2}{c}{\textbf{Bamboogle}}
& \multicolumn{2}{c}{\textbf{GAIA}}
 \\
\cmidrule(lr){3-4} \cmidrule(lr){5-6} 
& & F1 & LasJ & F1 & LasJ  \\
\midrule
\multirow{4}{*}{\textbf{Qwen-7B-SFT}}
& before training & 70.8 & 71.2 & 28.0 & 26.2  \\
& after training & 67.5 & 68.8 & 23.9 & 21.4 \\
& QwQ-32B & \textbf{74.5} & \textbf{76.8} & \textbf{39.3} & \textbf{36.9} \\
& GPT-4o-mini & 70.9 & \textbf{76.8} & 33.7 & 32.0  \\
\midrule
\multirow{3}{*}{\textbf{QwQ-32B-SFT}}
& before training & \textbf{83.5} & \textbf{88.0} & \textbf{48.9} & \textbf{50.5}  \\
& after training & 83.9 & 86.4 & 43.2 & 47.6  \\ 
& GPT-4o-mini & 80.0 & 80.8 & 40.5 & 44.7 \\
\bottomrule
\end{tabular}
}
\caption{Performance comparison across all benchmarks using different summarization models.}
\label{tab:summarization_model_selected_benchmarks}
\end{table}

\begin{table}[t]
    \centering
    \small
    \resizebox{1\linewidth}{!}{
        \begin{tabular}{l c c c c c c }
            \toprule
            \multirow{2.5}{*}{\textbf{Training Data}} 
            & \multicolumn{2}{c}{\textbf{Bamboogle}} 
            & \multicolumn{2}{c}{\textbf{GAIA}} 
            & \multicolumn{2}{c}{\textbf{AIME}} \\
            \cmidrule(r){2-3} \cmidrule(r){4-5} \cmidrule(r){6-7}
            & {F1} & {LasJ} 
            & {F1} & {LasJ} 
            & {F1} & {LasJ} \\
            \midrule
            - Reasoning   & 74.5 & 76.8 & \textbf{39.3} & 36.9 & 13.3 & 13.3  \\
            + Reasoning   & \textbf{76.9} & \textbf{80.8} & 37.2 & \textbf{37.9} & \textbf{20.0} & \textbf{20.0} \\
            \bottomrule
        \end{tabular}
    }
    \caption{Results of the SimpleDeepSearcher trained w/ and w/o reasoning data across three benchmarks.}
    \label{tab:mixed_reasoning_data_benchmarks}
\end{table}

\subsection{Effect of Summarization Model}
\label{sec:summary_model}

This part investigates the impact of the summarization model on overall performance. 
We fix the reasoning model and conduct a comparative analysis of overall performance using different summarization models. As shown in Table ~\ref{tab:summarization_model_selected_benchmarks}, QwQ-32B demonstrates the strongest summarization capability and is therefore selected as the summarization model for all reasoning models. Furthermore, using fine-tuned models for summarization leads to performance degradation on downstream tasks compared to their pre-trained counterparts. This might be attributed to the reduced long-text summarization ability of the fine-tuned models, due to the distributional shifts on a limited task and domain of the training data. This decline is more pronounced for models with fewer parameters.

\subsection{Effect of Additional Reasoning Data}



\begin{figure}[t]
    \centering
    \includegraphics[width=\linewidth]{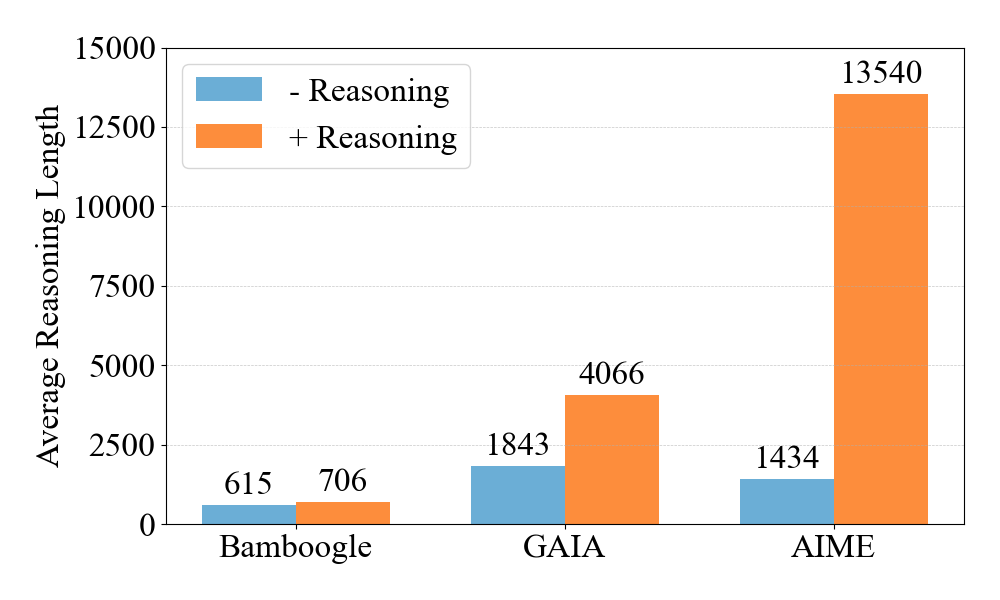}
    \caption{Average reasoning length across different benchmarks w/ and w/o reasoning data for training.}
    \label{fig:mixed_reasoning_data_length}
\end{figure}

We further investigate the impact of incorporating complex mathematical reasoning data on Qwen2.5-7B-Instruct. As shown in Table~\ref{tab:mixed_reasoning_data_benchmarks}, this leads to consistent performance gains across all benchmarks. Furthermore, Figure~\ref{fig:mixed_reasoning_data_length} and Table~\ref{tab:mixed_reasoning_data_search} reveals significant alterations in the model's behavioral patterns on two kinds of tasks: for tasks emphasizing complex reasoning (\eg AIME, GAIA), the model generates longer and more in-depth reasoning outputs; for search tasks (\eg Bamboogle), the model performs more searches and explores more thoroughly. These findings suggest that incorporating complex reasoning data helps the model learn to adapt its reasoning and search strategies to the specific demands of a task. This adaptability is critical for addressing complex and diverse queries.

\begin{table}[t!]
    \centering
    \small
        \begin{tabular}{l c c c}
            \toprule
              \multirow{2.5}{*}{\textbf{Training Data}}
            & \multicolumn{3}{c}{\textbf{Search Count}} \\
            \cmidrule(lr){2-4}
             & {Bamboogle} & {GAIA} & {AIME} \\
            \midrule
            
                 - Reasoning   & 1.552 & 1.757 & 0 \\
                 + Reasoning   & 1.672 & 1.845 & 0 \\
            \bottomrule
        \end{tabular}
    \caption{Average search count across different benchmarks of the model trained w/ and w/o reasoning data.}
    \label{tab:mixed_reasoning_data_search}
\end{table}

\section{Conclusion}
In this work, we present SimpleDeepSearcher, a lightweight yet effective framework for deepsearch tasks, addressing the limitations of existing RAG methods that rely heavily on complex training paradigms or suffer from distributional mismatches. By leveraging realistic web search simulations and a multi-criteria data curation strategy, we construct high-quality training trajectories that enable efficient supervised fine-tuning. Despite using only 871 curated samples, our method achieves substantial gains over RL-based baselines across diverse in-domain and out-of-domain benchmarks. Our results highlight the potential of strategic data engineering to empower deep search reasoning.

\section*{Limitation}

Despite our substantial efforts, this work is subject to two limitations stemming.
Due to limitations in training resources and hardware, we conducted distillation training on 7B and 32B models. In future work, we plan to train and evaluate our framework on larger-scale models (\ie 72B) to further verify its generalization capability and robustness. Additionally, because of the inherent difficulty in synthesizing multi-hop data, the original data used for distillation primarily consisted of relatively simple multi-hop questions. If more realistic and challenging multi-hop queries can be synthesized in the future, applying our framework for filtering and training may yield even better performance.

\section*{Ethics Statement}

We strictly adhere to ethical standards. We follow the relevant licenses and guidelines for dataset usage, ensuring that no personal or offensive information is included. AI assistance was only utilized during the paper refinement process. Our trained models do not display any potential biases or discriminatory behavior, and we rigorously comply with research ethics throughout the entire development and evaluation process.

\section*{Acknowledgments}

This work was partially supported by National Natural Science Foundation of China under Grant No. 92470205 and 62222215, Beijing Natural Science Foundation under Grant No. L233008 and Beijing Municipal Science and Technology Project under Grant No. Z231100010323009.

\bibliography{custom}

\appendix



\newpage
\appendix

\clearpage

\section{Related Work}
\paratitle{Retrieval-Augmented LLMs.}
To improve the factual precision of LLM-generated texts~\citep{rag_survey}, researchers enhance LLMs with retrieval-augmented generation (RAG)~\citep{guu2020retrieval}. Various approaches have been proposed, such as branching-based methods~\citep{kim2024sure}, summarization-based methods~\citep{li-etal-2023-compressing}, and adaptive retrieval techniques~\citep{jeong2024adaptive}.
With the increase in model parameters, LLMs have demonstrated chain-of-thought reasoning capabilities~\cite{wei2022chain}, and many researchers to integrated such reasoning with RAG via prompt engineering~\citep{shao2023enhancing, trivedi2023interleaving}. 
Other studies have attempted to distill retrieval abilities into smaller models through supervised fine-tuning~\cite{self-rag}.
However, these approaches limit the model's capacity with a fixed reasoning path.

\paratitle{Enhancing LLMs with Search.} 
Recently, several deep search frameworks are proposed~\citep{alzubi2025open}. They integrate large language models with search engines in a more flexible and dynamic manner.
Search-o1~\citep{search_o1} simulates deep search in LLMs through prompt engineering, allowing them to retrieve information independently during multi-step reasoning.
R1-Searcher~\citep{r1-searcher} and Search-R1~\citep{search-r1} equip large language models with retrieval tools and train them end-to-end using reinforcement learning.
This approach effectively enhances the model’s ability to interleave reasoning with retrieval during inference.
However, due to the inherent complexity of RL and its high computational demands, conducting large-scale experiments on full-sized LLMs remains challenging. 
SimpleDeepSearcher synthesizes high-quality training data via broad query sampling and precise filtering, enabling strong deep search performance with minimal training cost.

\section{Details of Diversity-Aware Query Sampling}
\label{sec:daqs}



\begin{figure}[t]
    \centering
    \includegraphics[width=1\linewidth]{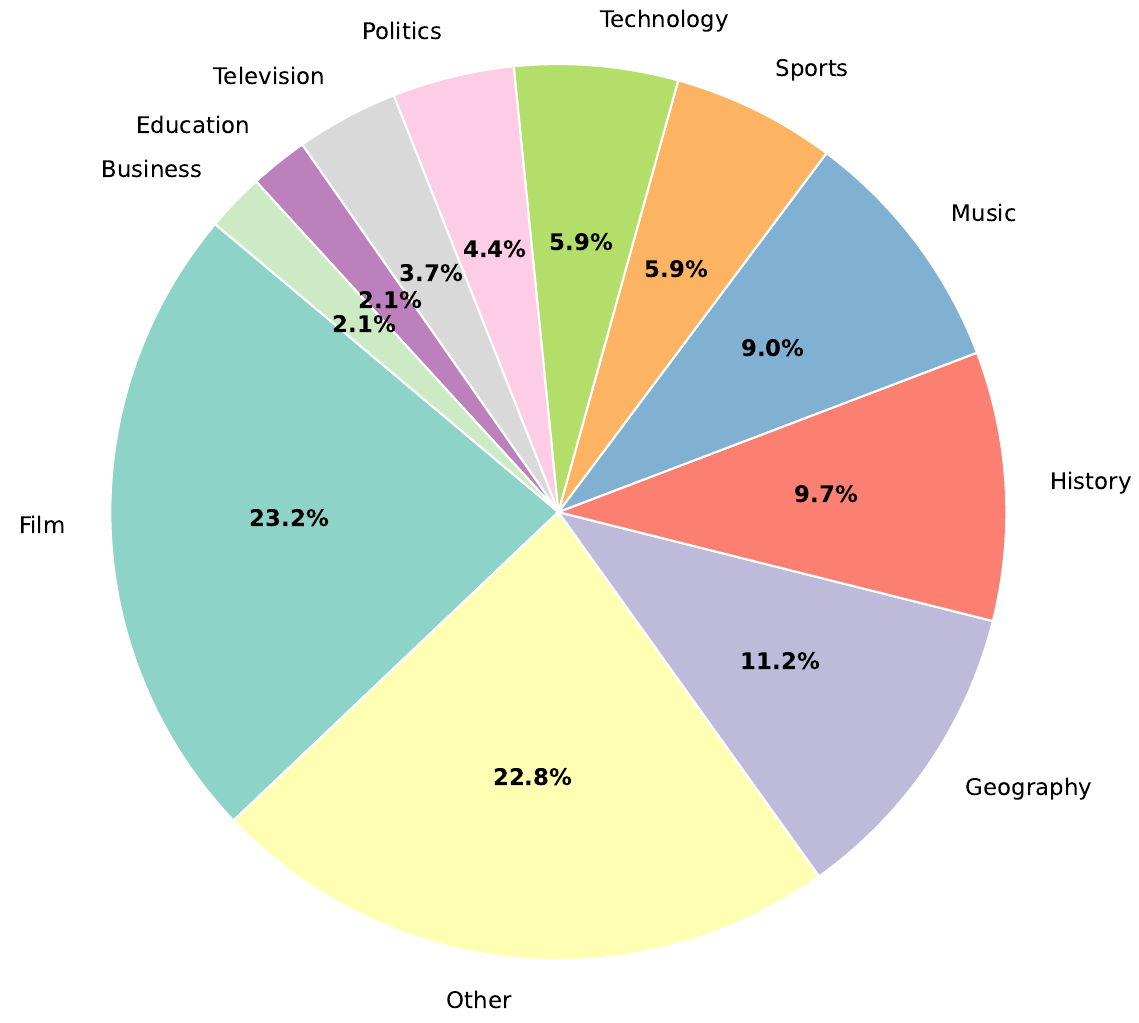}
    \caption{Domain distribution of the data before filtering.}
    \label{fig:domain_distribution}
\end{figure}

In analyzing open-source data, we identified three critical limitations:

(1) Domain-specific overrepresentation creating skewed knowledge distributions.
As shown in the Figure~\ref{fig:domain_distribution}, we present the domain distribution of the pre-filtered data. It can be observed that certain domains (such as film and geography) account for a considerable proportion. This imbalance risks inducing uneven knowledge distributions in the training data.

(2) Repetitive syntactic patterns reducing linguistic variability.
Due to the construction methods of open-source datasets, we observe substantial redundancy in syntactic structures. A typical example is the prevalence of ``A and B'' style comparative queries (\eg ``Do both directors of films Paper Bullets and Karakolda Ayna Var share the same nationality?'' vs. ``Do both directors of films Jatt Juliet and Sciopèn share the same nationality?''). Similarly, numerous queries repetitively compare identical attributes such as age.

(3) Semantic simplicity thresholds below real-world query complexity.
Many queries in open-source datasets are overly simplistic, such as ``What nationality is John Harbaugh’s father?''. Such questions impose only minimal demands on deep search or reasoning, as they can often be answered through a single lookup. Consequently, their utility in fostering more advanced model capabilities is limited.

We define ``core semantic constituents'' as follows:

$\bullet$ Key entities (\eg films, people, locations)

$\bullet$ Critical attributes (\eg age, duration, population)

$\bullet$ Core relationships (\eg comparison, causality)

$\bullet$ Measurement dimensions (\eg time, quantity)

For example, in the query ``Which film whose director is younger, Charge It To Me or Danger: Diabolik?'', the extracted keywords based on the above schema are ``film'' and ``age''.

\section{DPO Detailed Settings}
\label{sec:dpo_setting}

Our objective was to identify answer trajectories that were both correct and demonstrated efficient reasoning and search paths. To this end, we construct preference pairs $(R_w, R_l)$, where $R_w$ denotes the preferred trajectory and $R_l$ the rejected one. We repurposed our previously established pipeline for query sampling and data synthesis. During the data synthesis stage, we generate responses using the strongest SFT-trained model, SDS-QwQ-32B-SFT, and the target model to be optimized, SDS-Qwen-7B-SFT. Responses generated by SDS-QwQ-32B-SFT that pass both the formatting and reasoning path control checks are treated as positive examples, while those generated by SDS-Qwen-7B-SFT that fail these checks are treated as negative examples. Ultimately, we construct a dataset consisting of approximately 875 training pairs. 


For Direct Preference Optimization (DPO) training, we utilize a learning rate of $5 \times 10^{-7}$, a $\beta$ of $0.1$, training for $5$ epochs with a batch size of $256$, a warm-up ratio of $0.1$, and a maximum sequence length of $10000$.



\section{REINFORCE++ Detailed Settings}\label{sec:appendix-on-policy}

\begin{figure}[t]
    \centering
    \includegraphics[width=1\linewidth]{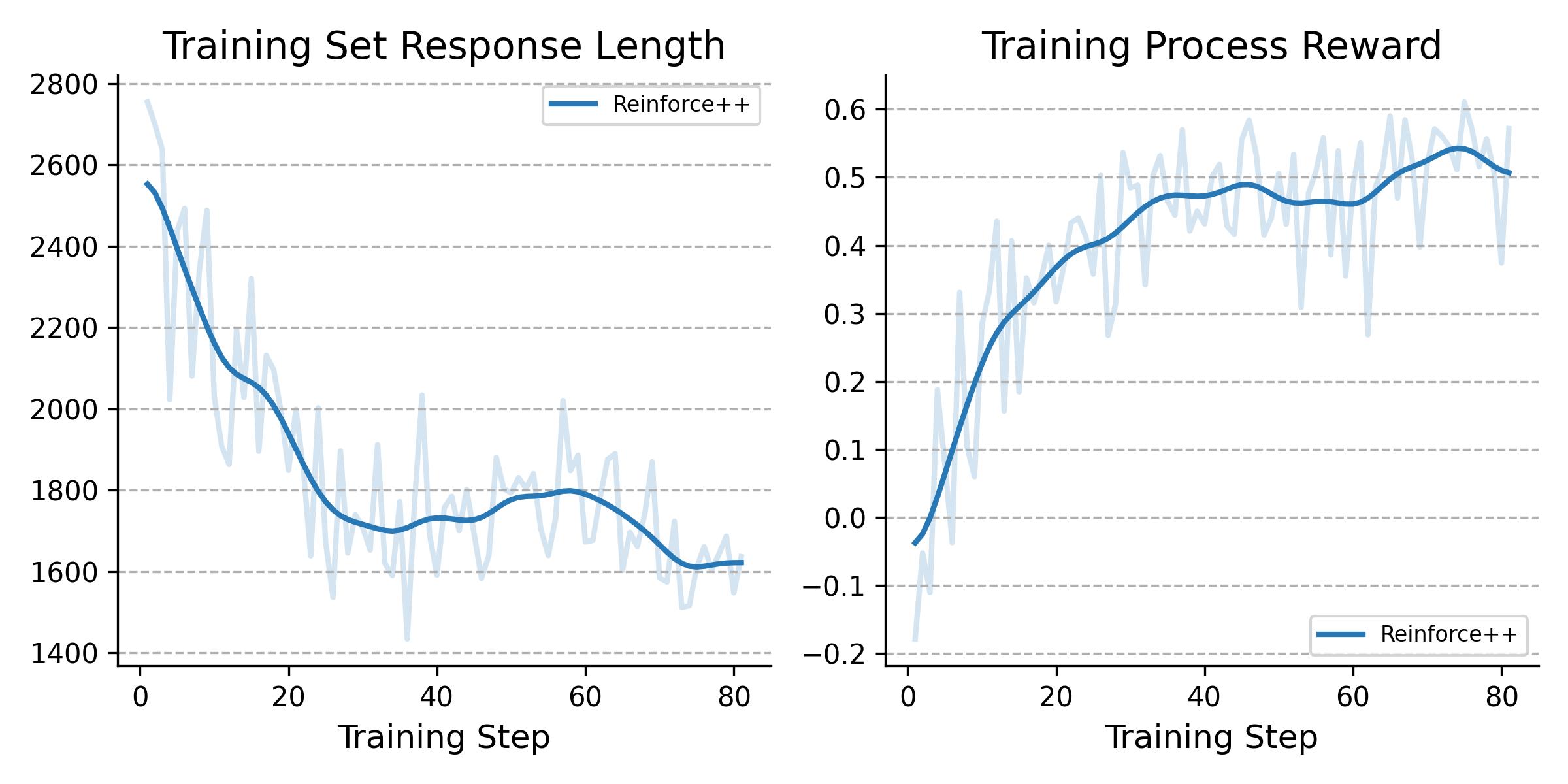}
    \caption{Changes in Sequence Length and Reward During REINFORCE++ Training.}
    \label{fig:rl_log}
\end{figure}

To construct the reinforcement learning (RL) dataset, we utilized the model that had been trained though SimpleDeepSearcher to perform rollout sampling on the training sets of 2Wiki and HotpotQA. For each question, eight candidate responses were generated. From this pool, we selected 2480 samples corresponding to questions with one to six correct answers, ensuring diversity in the RL training data.

The reward function employed in REINFORCE++ consists of two components: an answer reward and a format penalty. The answer reward is calculated as the F1 score between the predicted answer and the reference answer, providing a direct measure of response accuracy. In addition, a discrete format penalty of $-2$ is applied if any of the following undesirable behaviors are detected: 

$\bullet$ \textit{Self-Retrieved Content:} The model fabricates content that is not retrieved from external sources.

$\bullet$ \textit{Contains Gibberish:} The generated output contains nonsensical, irrelevant, or corrupted text segments.

$\bullet$ \textit{Excessive Analytical Markers:} The response contains more than 5 occurrences of phrases such as \textit{Alternatively}, \textit{Wait}, or \textit{Hmm}, which are treated as signals of incoherent reasoning.

$\bullet$ \textit{Lack of Boxed Answers or Excessive Reasoning Length:} The model either executes more than 8 retrieval steps or the token length of the analytical content between any two retrievals exceeds 8,096 tokens.

If none of these conditions are met, no penalty is applied.
To maintain on-policy training throughout the RL process, we adjusted the batch size to ensure that learning was based on the most recent policy rollouts. Figure~\ref{fig:rl_log} shows the variations in response length and reward values observed during the training process.

\section{Model Performance Enhancement Analysis Settings}
\label{sec:model_enhancement}
We conduct a comparative analysis of Qwe2.5-7B-Instruct and QwQ-32B before and after training across the 2Wiki, Bamboogle, and MuSiQue benchmarks. During inference, we fix the summarization model to QwQ-32B across all comparisons to eliminate potential interference from the summarization component (the impact of the summarization model will be further discussed in Section~\ref{sec:summary_model}).


\section{Instruction Templates}
\label{sec:instrcution_templates}

\begin{figure*}[t]
    \centering
    \includegraphics[width=1\linewidth]{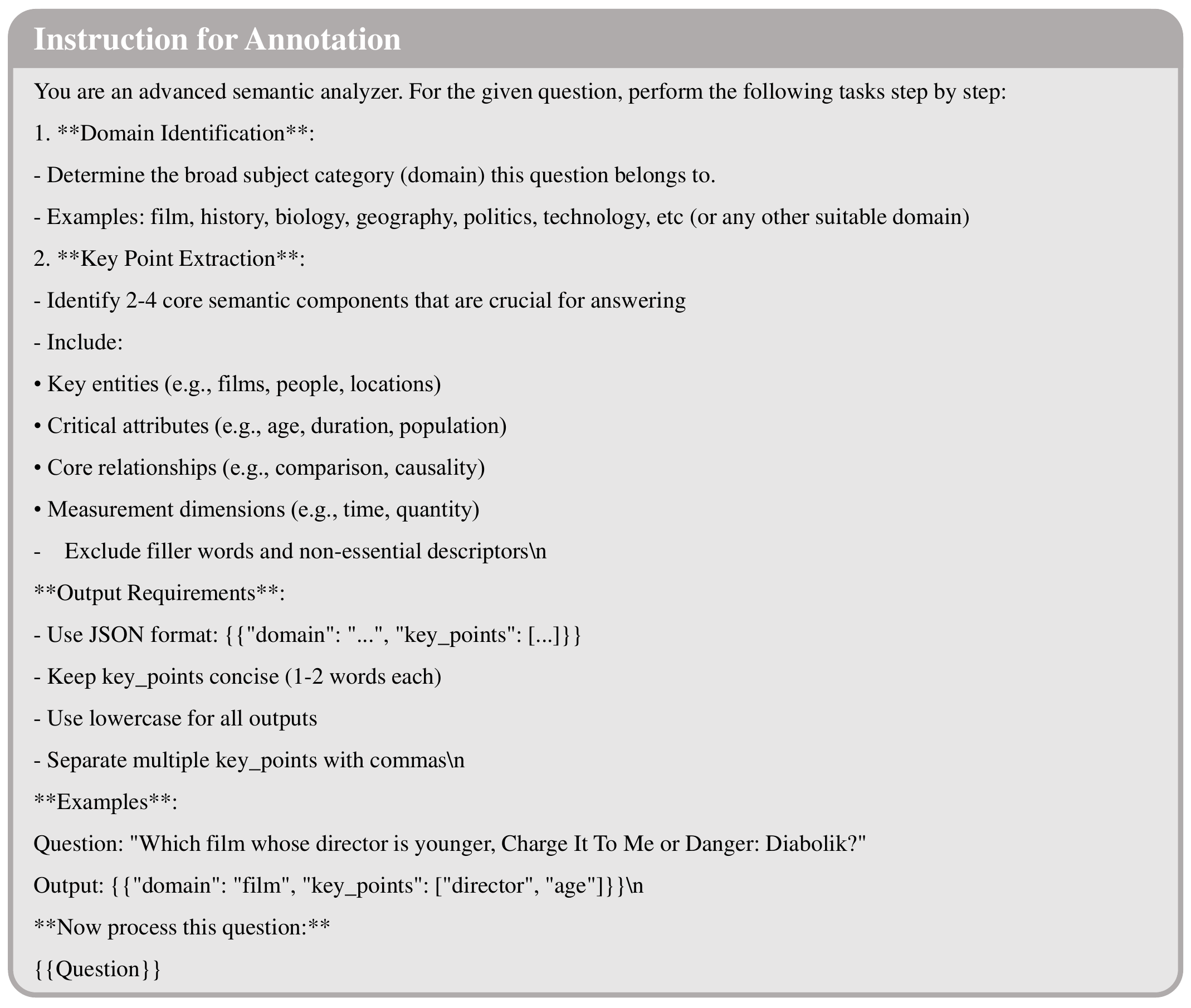}
    \label{fig:template_annotation}
\end{figure*}

\begin{figure*}[t]
    \centering
    \includegraphics[width=1\linewidth]{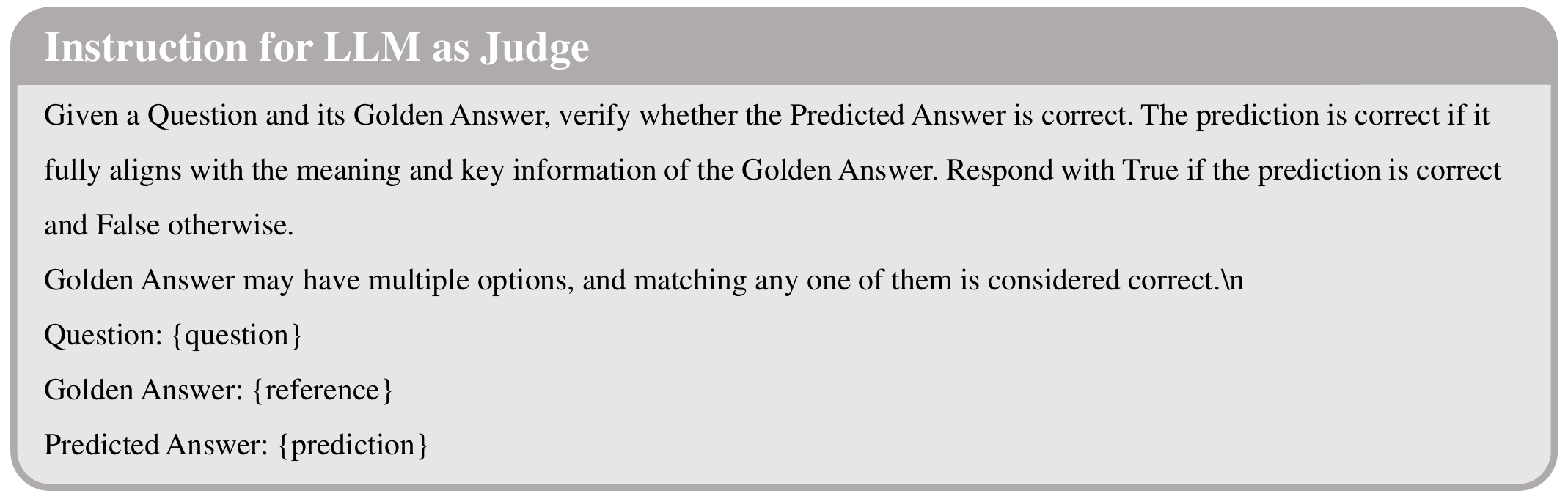}
    \label{fig:template_judge}
\end{figure*}

\begin{figure*}[t]
    \centering
    \includegraphics[width=1\linewidth]{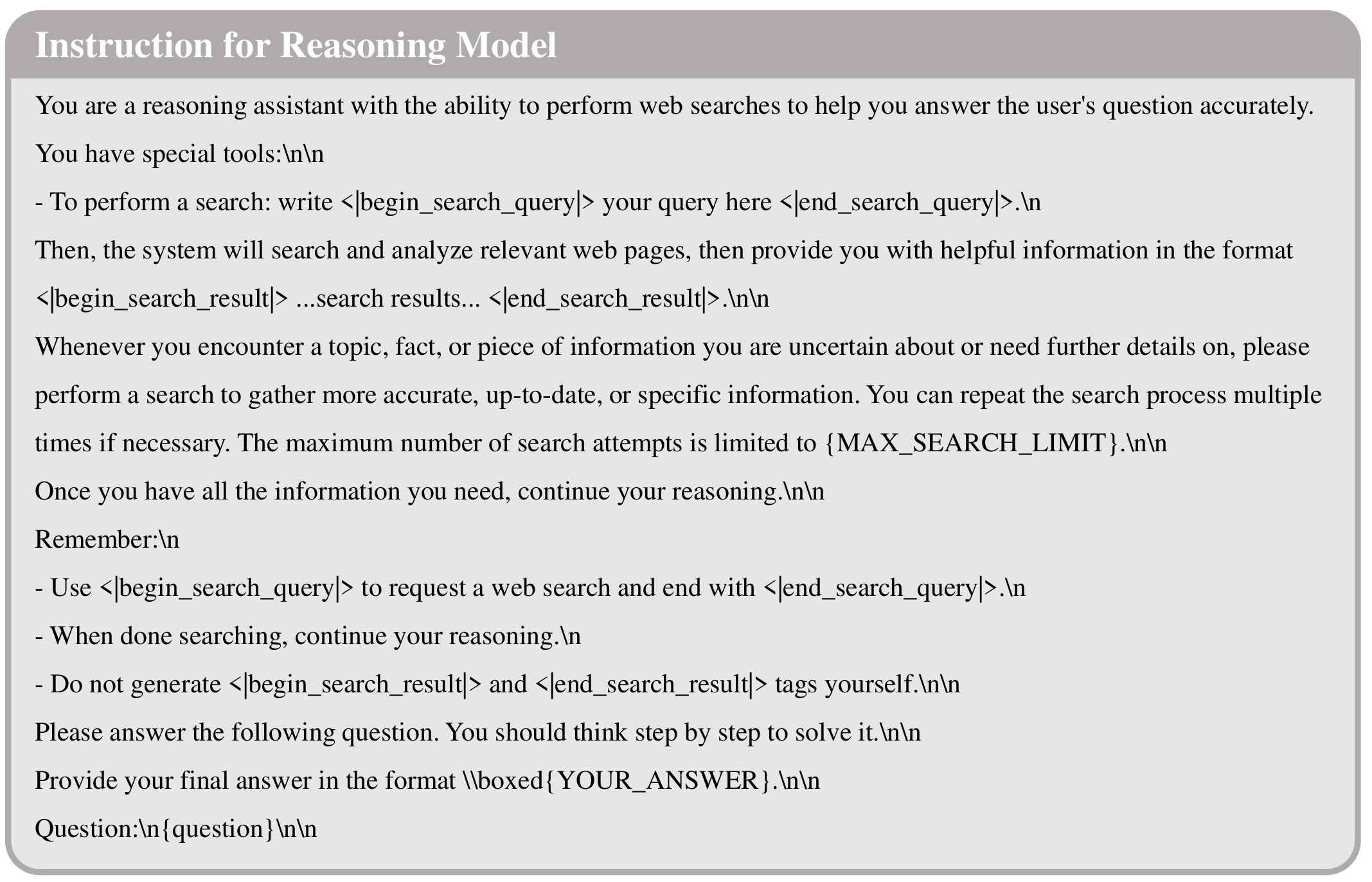}
    \label{fig:template_reasoning}
\end{figure*}

\begin{figure*}[t]
    \centering
    \includegraphics[width=1\linewidth]{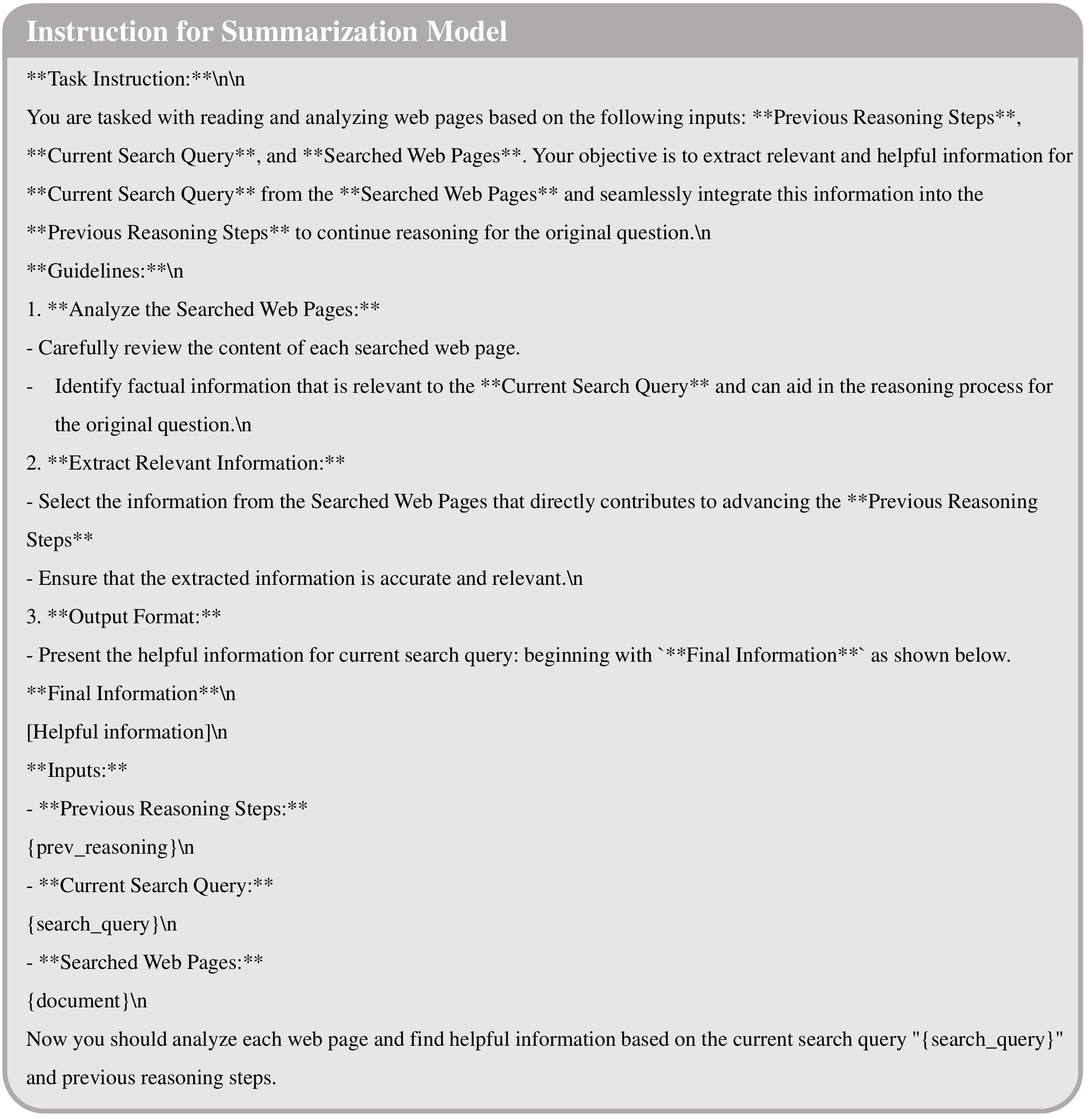}
    \label{fig:template_summary}
\end{figure*}
\end{document}